\documentclass{article}
\usepackage{spconf,amsmath,graphicx}

\title{Conformer LLMs - Convolution Augmented Large Language Models}
\name{Prateek Verma}
\address{
Stanford University\\
353, Jane Stanford Way,\\
Stanford, California, 94305
}

\begin{document}
\maketitle
\begin{abstract}
This work builds together two popular blocks of neural architecture, namely convolutional layers and Transformers, for large language models (LLMs). Non-causal conformers are used ubiquitously in automatic speech recognition. This work aims to adapt these architectures in a causal setup for training LLMs. Transformers decoders effectively capture long-range dependencies over several modalities and form a core backbone of modern advancements in machine learning. Convolutional architectures have been popular in extracting features in domains such as raw 1-D signals, speech, and images, to name a few. In this paper, by combining local and global dependencies over latent representations using causal convolutional filters and Transformer, we achieve significant gains in performance. This work showcases a robust speech architecture that can be integrated and adapted in a causal setup beyond speech applications for large-scale language modeling.

\end{abstract}
\section{Introduction and Related Work}
\label{sec:intro}
Large Langauge Models have demonstrated significant capabilities across modalities, ushering in a new revolution and frantic interest in artificial intelligence. They have developed super-human abilities in dialogue systems, speech recognition, and image processing. They are built on a core building block of Transformer architectures \cite{vaswani2017attention}, which have been used not only for LLMs but also for understanding audio \cite{verma2021audio}, speech recognition \cite{schneider2019wav2vec}, image understanding \cite{dosovitskiy2020image}, text \cite{brown2020language}, protein sequences \cite{madani2020progen}, robotics \cite{shridhar2023perceiver} to name a few. They operate on a causality assumption, and by using a simple proxy goal of prediction of a text token, they can capture and model the characteristics of the input context and summarize it. Recent advancements in large language models have also trickled down to real-world setups such as reinforcement learning \cite{chen2021decision}, step-by-step reasoning \cite{wei2022chain}, building dialogue agents \cite{glaese2022improving}, solving math problems \cite{lewkowycz2022solving} as well as coding \cite{chen2021evaluating}. The ability to do one-shot reasoning by choosing a prompt from another modality has given these architectures nothing short of magical properties, as shown in the paper ``Socrates architectures" \cite{zeng2022socratic}. They also exhibit emergent properties that develop by scaling the number of parameters \cite{weiemergent}. They have also been able to learn multi-modal representations by solving proxy tasks of next token prediction \cite{driess2023palm}, which previously would have required a complex pipeline to combine two modalities, e.g., text and audio as shown in \cite{haque2019audio}. 
\begin{figure*}[t]
  \centering
  \hspace*{8.8pt}
  \includegraphics[width=\linewidth, height=6cm]{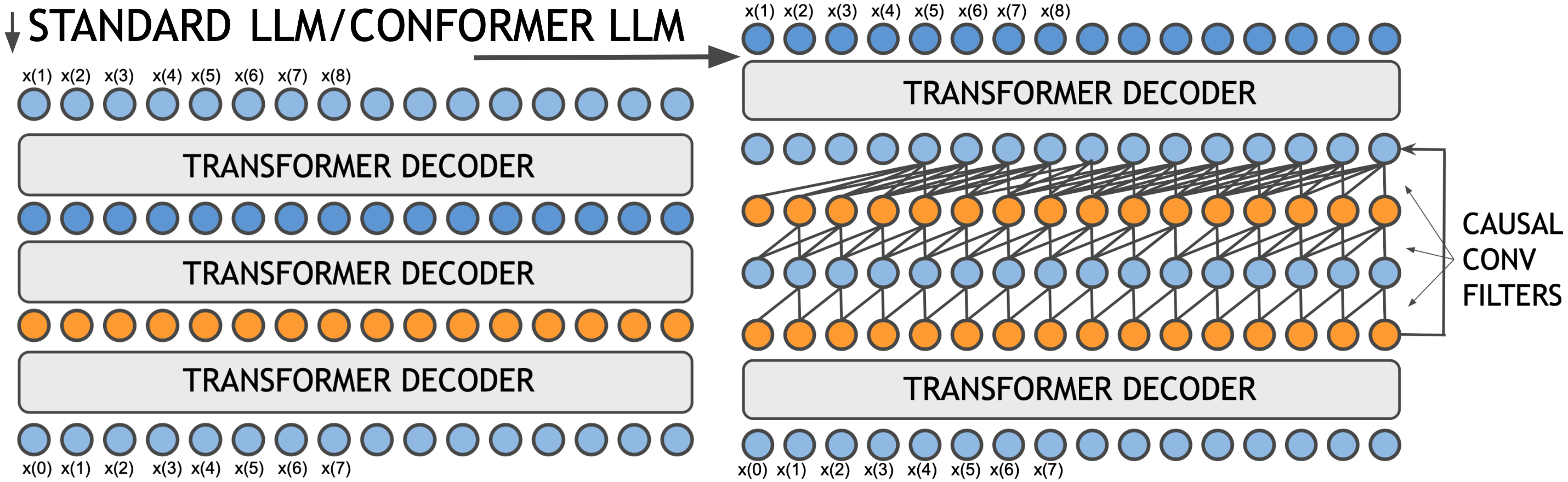}
  \caption{We compare a Transformer based decoder langauge model (left) vs. that of a Conformer langauge Model (right). In between \textit{every} decoder block in the model, we sandwich a set of causal convolutional layers with small kernel sizes.}
  \label{fig:speech_production}
\end{figure*} 
Just before the advent of Transformer architectures, we saw a brief shift from a traditional LSTM-based architecture \cite{45610} to modeling sequences to that of one that uses dilated convolution for, e.g., wavenet \cite{oord2016wavenet}. 
They were also explored for natural language processing applications in ByteNet, showing gains over traditional LSTM architectures \cite{kalchbrenner2016neural}. This shift to dilated convolutions resulted in several increased gains for the task of time-series prediction in various fields apart from raw audio as described in \cite{verma2021generative} and natural language processing\cite{brown2020language}. However, the main drawback of this shift was that the topology of how the convolutional filters operated is fixed. In contrast, an attention mechanism could allow depending on the dataset, to decide the topology automatically. However, with the advent of Transformers, slowly convolutional-based architectures were phased out, and end-to-end architectures were trained for causal setups such as language modeling and non-causal setups such as in audio \cite{verma2020framework} or vision\cite{wu2021cvt}. Conformers \cite{gulati2020conformer}, first proposed for ASR, combined convolutional filters with self-attention and feed-forward layers in Transformers. It gave significant gains for speech recognition by utilizing the best of both worlds. Convolutional augmented image transformers similar in non-causal setups have shown similar gains for understanding images and visual content \cite{wu2021cvt}. Conformers had its origins from CLDNN  architectures \cite{sainath2015convolutional} by replacing LSTM layers with Transformers in the sense that convolutional layers are good feature extractors, LSTMs are good modeling sequences, and MLP layers transform the learned features to more separable space. In this work, we utilize conformers to train large language models in a causal setup. All of our convolutional filters are causal. This architecture thus allows the model to have local and global connections while learning kernels that can filter out/understand dependencies according to the task. One can achieve significant gains by designing hand-made filters in a non-causal setup which was explicitly done by \cite{tamkin2020language}. Our proposed architecture outperforms a purely Transformer based architecture by adding small shallow context convolutional kernels that are causal and which incorporate the best of both worlds. We report results on two standard datasets: For language modeling over text characters and raw audio. For this paper, we do not achieve the state of the art as the number of parameters we use is far smaller than a GPT-2/3. However more importantly, we show experiments showing increasing our systems' complexity and gains achieved with/without using causal convolutional on every intermediate embedding sandwiched between transformer layers. The contributions of the paper are as follows: i) We showcase gains in using a shallow layer of convolutional filters sandwiched in between two transformer decoder modules and achieve significant gain training of large language models; ii) We show that a Conformer LLM, i.e., a convolution augmented LLM scales well similar to a traditional large language model with embedding size, number of heads and scaling of a convolutional block, which augurs well with integrating these architectures with a decoder only blocks. The paper's organization is as follows: Section 2 describes the dataset we used for our experiments and explains the conformer architecture. Our experimental setup follows this section and shows the gains we achieve using convolution-augmented language models.
\section{Dataset}
\label{sec:dataset}
To showcase the strength of our paper, we run experiments for two datasets. The goal in a specific dataset is simple: given the context of previous samples/tokens/characters, predict the next token from the context. For raw waveforms, we make use of the YouTubeMix dataset. It consists of piano sounds at 8KHz, 8-bit quantized signal totaling about 4 hours of audio files. This can be treated as a long-time series of discrete values ranging from 0-255. We use YouTube Mix dataset as reported previously in \cite{verma2022goodbye} and \cite{goel2022s} with the same training/validation/test splits. In all these experiments, the context is fixed to 256 tokens/samples/events. For the text data, we utilize the text8 dataset \cite{text8} that consists of predicting 27 tokens, which are 26 lowercase alphabets and a single token depicting space. 

\section{CONFORMER LANGUAGE MODEL}

This section describes various blocks of our proposed architecture; concisely, we use causal convolutional filters on the intermediate embeddings after every Transformer decoder layer. However, for the sake of completeness, and this paper, we will explain the entire architecture in detail.

A Transformer decoder architecture is typically used in langauge modeling tasks. It differs from Transformer Encoder in that we use a causal mask so that every layer of intermediate/final transformer embeddings is only a function of the past samples/tokens/embeddings. 

\subsection{Feed Forward and Convolutional Module} We use a feed-forward architecture consisting of a single layer of hidden units two times the dimension of the embeddings, in our case fixed as 128. This is used inside the Transformer block as proposed by \cite{vaswani2017attention}. However, we use causal convolutions on intermediate embeddings for each decoder block. Each convolution model has two variants: The small variant consists of learning causal filters across the embedding dimension E with kernel widths of 3 and 7 across the token dimension. We learn the total number of filters to be $E$ in all layers. This convolutional output is then added back to the input of the convolutional module (which, to begin with, was the output of the previous layer of the Transformer) which is a skip connection enabling faster convergence.  However, to not impose any inductive biases in the structure of the dependencies, in the last Transformer decoder layer, we do not use it before it. For the large convolutional block, we use filters of kernel width, namely of {2,3,5} with the number of filters being {$2E,2E, E$}. Each of the convolutional filter outputs is passed through a relu non-linearity. As described before, we use a skip connection, similar to the core Transformer architecture. Hence the input to the 2-layer convolution block is then added back to the output of the convolutional layers. 
\section{Experiments and Results}
\label{sec:exp}

\subsection{Baseline} For all three datasets, we fix a baseline architecture as follows: We use six layers with ten attention heads, with a context of 256 tokens, both raw waveform and natural language characters. The size of the embedding $E$ for the Transformer is fixed to be 128. The hidden dimension is 512. All of the models are trained from scratch using Tensorflow framework \cite{abadi2016tensorflow}. 
\subsection{Performance on modalities}

In the subsequent experiment, we use text8 and youtube-mix dataset to investigate scaling with the number of heads, embedding dimension, and scaling of the convolutional block. We keep the training schedules, loss function to train, and training set up the same except for the modifications mentioned in the subsequent sections. For all the experiments, we train for 30 epochs, with a learning rate of 3e-4 for ten epochs, and then 1e-4 for the next ten epochs, followed by 1e-5 for the last five epochs, with a dropout rate of 0.1 for the feed-forward blocks. This configuration acted as our baseline architecture. For the conformers baseline, we add two layers of causal convolution between every layer of the Transformer decoder except for the final layer. These two convolutional layers had kernel sizes 3 and 7, respectively, with 128 filters, which was the same as the size of the embedding dimension. We added a relu non-linearity after each of the convolutional filters. Table 1 below showcases the results of our experiments. The gain of a NLL of about 0.04 for text and for piano a change of 0.17 is quite significant improvement, given that we do not add a significant number of parameters.

\begin{table}[ht]
  \caption{Comparison of Negative Log-Likelihood Loss (NLL) and Test Accuracy for Text-8 and Youtube-Mix dataset \\}
	\centering
	\begin{tabular}{|c|c|c|}
		\hline
		Model + Dataset  & \# Params &NLL score\\\hline
		Text-Baseline LLM & 5.58M  & 1.03\\
        \textbf{Text-Conformer LLM} & 5.61M & 0.99\\
        \hline
        \hline
        Piano-Baseline LLM & 5.58M & 2.58\\
        \textbf{Piano-Baseline LLM} & 5.61M  & 2.41\\
		 \hline
		
	\end{tabular}
	\label{tab:example}
\end{table}

\subsection{Scaling with the number of heads} In this section, we explore how the performance of conformer LLMs scales with the number of heads. For this, we use the same context as before, i.e., given a 256-length context, we predict the next token. The convolutional block of the conformer is fixed with two conv layers, each causal filter, with stride one, and the number of filters equal to the embedding dimension of the Transformer block, with kernel size as 3 and 7, with skip connection. There are six conformer blocks, with a feed-forward dimension of 128 and embedding size of 64, kept constant for all of the heads chosen, namely, 4, 8, 16, and 32, respectively. The results are shown in the Table below. We see that the conformer block, similar to a Transformer block, scales with the number of attention heads.

\begin{table}[ht]
  \caption{Comparison of Negative Log-Likelihood Loss (NLL) and Test Accuracy for Attention Heads \\}
	\centering
	\begin{tabular}{|c|c|c|}
		\hline
		\# of Attention Heads  & \# Accuracy &NLL score\\\hline
		4 Heads & 64\%  & 1.16\\
        8 Heads & 65.5\% & 1.14\\
        16 Heads & 65.1\% & 1.12\\
        \textbf{32 Heads} &  \textbf{65.8\%}  & \textbf{1.10}\\
		 \hline
		
	\end{tabular}
	\label{tab:example}
\end{table}

\subsection{Scaling with Embedding Dimension}
In this experiment, we explore how our architecture scales with the size of the embedding dimension. We expect the model to scale in performance with the embedding size in standard Transformer modules. We expect the same to hold for the Conformer module. We keep the base conformer module the same: We have 6 Transformer layers with eight attention heads in each layer, with a convolutional module in between each Transformer block: consisting of 2 causal convolutional layers with kernel sizes 3 and 7, layers with skip connection, and the number of filters equal to the embedding dimension chosen. We experiment with three dimensions of E, as shown in the Table 3 below. It confirms that our architecture again scales well with the embedding size.

\begin{table}[ht]
  \caption{Comparison of Negative Log-Likelihood Loss (NLL) and Test Accuracy for Embedding dimension \\}
	\centering
	\begin{tabular}{|c|c|c|}
		\hline
		Embedding Dimension  & \# Accuracy &NLL score\\\hline
		16 & 53.6\%  & 1.16\\
        64  & 65.5\% & 1.14\\
        \textbf{256} & \textbf{70.2\%} & \textbf{0.96}\\
		 \hline
		
	\end{tabular}
	\label{tab:example}
\end{table}

\subsection{Scaling of Convolution block} In this experiment, we see what can be the effects of a more robust convolutional block that augments the intermediate Transformer embeddings. To reiterate again, we keep the convolutional blocks as causal for all of the convolutional layers. Here we again show how our architecture does with scaling. We keep the parameters the same as before for the baseline convolutional block. The core Transformer backbone consists of 6 layers with an embedding dimension of 256, 8 heads, and a feed-forward dimension twice the embedding dimension. We keep the parameters for the small convolutional module as follows: We have convolutional modules consisting of causal convolutional filters after every layer of the Transformer. For the small model, we have two layers: each consisting of 256 filters of kernel size 3 and 7, respectively. We add two more convolutional layers of kernel sizes 2,3,5, and 7 for the larger model. The goal of increasing the complexity is to learn and impose more constraints by using convolutional operations as feature extractors. We see significant improvements in likelihood scores and accurate prediction of the next token, as shown in the Table 4. 
\begin{table}[ht]
  \caption{Comparison of Negative Log-Likelihood Loss (NLL) and Test Accuracy for Size of the Convolutional block \\}
	\centering
	\begin{tabular}{|c|c|c|}
		\hline
		\# Size of Convolutional Block  & \# Accuracy &NLL score\\\hline
		Small  & 70.15\%  & 0.965\\
        \textbf{Large} & \textbf{70.25\%} & \textbf{0.958}\\
		 \hline
		
	\end{tabular}
	\label{tab:example}
\end{table}

\label{sec:res}

\section{Conclusion and Future Work}

We have showcased the powerfulness of a convolutional augmented Transformer for the case of language modeling. We see that by adding a small number of convolutional parameters or, in other words, augmenting the Transformers with convolutional layers, we achieve significant gain in performance. We apply a causal convolutional block to the intermediate embeddings of every Transformer layer that can learn the best of two worlds: Transformers understand dependencies over long time scales, and convolutional filters act on those embeddings to transform them to a more separable space. This, similar to CLDNN or Conformer architectures, brings together three fundamental blocks of neural network advancements: attention, fully connected architecture, and convolutional layers. Through ablation studies, we show how our architecture works for two modalities and achieves gains in natural language processing, raw audio. Our architecture scales with the number of attention heads, parameters of convolutional blocks, and size of the intermediate embeddings, as shown in the experiment. The improved performance of conformers augers well for achieving faster convergence or performance gains with a similar number of parameters as compared to not adding a few parameters of convolutional block. There are several exciting ways of further exploring how we can combine these blocks, and it is an interesting future direction to improve the performance.

\section{Acknowledgements}
This work was supported by the Stanford Institute of Human-Centered AI (HAI) through a Google Cloud computing grant. This paper originated while working on genomic sequences in Kundaje Lab at Stanford University. I want to thank Prof. Kundaje and his group for providing a creative environment that led to this work.

\bibliographystyle{IEEEbib}
\bibliography{refs}

\end{document}